\documentclass[a4paper,UKenglish,cleveref, autoref, thm-restate, pdfa]{lipics-v2021}
\pdfoutput=1

\hideLIPIcs

\nolinenumbers

\usepackage{tcolorbox}
\usepackage[ruled]{algorithm2e}

\title{A Constraint Programming Approach for \texorpdfstring{$n$}{n}-Day Lookahead Playoff Clinching in the NHL}
\titlerunning{CP Approach for n-Day Lookahead Playoff Clinching in the NHL}

\author{Gili Rosenberg}{Amazon Advanced Solutions Lab, Seattle, WA, 98170, USA
}{gilir@amazon.com}{https://orcid.org/0009-0007-6866-9949}{}
\author{Kyle E. C. Booth}{Amazon Advanced Solutions Lab, Seattle, WA, 98170, USA
}{kybooth@amazon.com}{https://orcid.org/0000-0001-6929-8042}{}
\author{J. Kyle Brubaker\footnote{Affiliated with Amazon at time of contributions.}}{Amazon Advanced Solutions Lab, Seattle, WA, 98170, USA
}{j.kylebrubaker@gmail.com}{https://orcid.org/0000-0002-6439-5270}{}
\author{Ruben S. Andrist}{Amazon Advanced Solutions Lab, Seattle, WA, 98170, USA
}{randrist@amazon.ch}{https://orcid.org/0000-0003-1126-6931}{}
\authorrunning{G. Rosenberg, K.\,E.\,C. Booth, J.\,K. Brubaker, and R.\,S. Andrist}

\Copyright{Gili Rosenberg, Kyle E. C. Booth, J. Kyle Brubaker, and Ruben S. Andrist}



\acknowledgements{The authors would like to acknowledge Helmut Katzgraber and Nino DiPasquale for their support, and Greg Inglis, Neil Pierson, Ben Resnick, and Katie Yates for productive discussions. }

\SeriesVolume{379}
\ArticleNo{13}

\begin{document}

\maketitle

\begin{abstract}
  In professional sports, a team has clinched the playoffs if they are guaranteed a postseason spot, regardless of the outcomes of any remaining games. As the season progresses, sports fans and other stakeholders are interested in precisely when, and under what conditions, their team will clinch the playoffs. In this paper, we investigate playoff clinching in the context of the National Hockey League (NHL), where it is computationally challenging to produce clinching scenarios due, in part, to complex tie-breakers. We present an algorithm that determines under which combinations of game outcomes in the next $n$ days a team will clinch the playoffs (i.e., ``$n$-day lookahead clinching''). Our approach is a custom tree search which employs various preprocessing techniques, pruning strategies, and node ordering heuristics to efficiently explore the space of possible outcomes. The tree search leverages a constraint programming (CP)-based subroutine for inference that determines if a team has clinched the playoffs  for some snapshot in time of the regular season (i.e., ``0-day lookahead clinching''). This CP subroutine aims to find a counter-example in which the team being evaluated is eliminated, taking into account qualification rules and the NHL's extensive list of tie-breakers. We validate the efficacy of our algorithm using hundreds of scenarios based on public NHL
  data for the seasons 2021-22 through 2024-25. The methods introduced can be readily extended to other metrics of interest, including mathematical proof of playoff elimination, clinching the President's Trophy, as well as clinching (or being eliminated from clinching) any other seed in the standings.
\end{abstract}

\section{Introduction}

The National Hockey League (NHL) is a professional ice hockey league with 32 teams participating from major cities across the USA and Canada. The NHL regular season has these teams competing against each other for one of a limited number of post-season spots, awarded based on the team's regular season performance. As the season progresses, NHL fans, marketers, and other stakeholders are interested in precisely when, and under which conditions, their team has \emph{clinched} the playoffs. A team has clinched the playoffs if they are guaranteed to make the postseason regardless of the outcomes of any remaining games. 

During the critical stages of the regular season (usually starting around March), the NHL begins publishing daily playoff clinch scenarios for teams affected by the games occurring later that evening (i.e., using a 1-day lookahead). As the league's tie-breaking rules have become more complex, determining clinching scenarios manually has become increasingly time-consuming and error-prone. Our work contributes an automated and mathematically rigorous alternative that can determine clinching scenarios fast enough for daily consumption. 

Over the years, the NHL has undergone many changes to the league's structure (teams, divisions, etc.), tie-breakers, rules for qualification, number of games, and so on.
This work is based on the state of the league's rules since the most recent change to playoff qualification and tie-breaking in the 2019-20 season (and is flexible to changes going forward).

The NHL regular season typically spans from October to April, with each team
playing 82 games. The league is divided into two conferences (Eastern and
Western), each further split into two divisions (Eastern: Atlantic and Metropolitan, Western: Central and Pacific). This structure plays a
crucial role in playoff qualification. At the end of the regular season, 16
teams qualify for the playoffs (8 from each conference). In each conference,
the top three teams from each division qualify, along with two wild card teams. This qualification structure introduces intricate scenarios where
teams compete not only within their divisions but also across their
conferences for wild card spots.

Each NHL game must produce a winner. If the score is tied after  regulation, the game proceeds to an overtime period, and if still tied, a shootout. Therefore, NHL games have six possible outcomes from the perspective of a team playing a game -- regulation win (RW), overtime win (OTW), shootout win (SOW), shootout loss (SOL), overtime loss (OTL), and regulation loss (RL). A win in any fashion awards 2 points, an overtime or shootout loss awards 1 point, and a regulation loss awards 0 points. In the NHL, teams are ranked first by points. When teams are tied in points, a series of
tie-breakers \cite{nhl_tiebreakers} is applied, see \cref{tab:nhl-tiebreakers}.

\begin{table}[htbp]
\small
\centering
\begin{tabular}{cl}
\hline
\textbf{Tie-breaker} & \textbf{Criterion} \\
\hline
(1) & Point percentage \\
(2) & Regulation wins \\
(3) & Regulation wins + Overtime wins \\
(4) & Total wins \\
(5) & Head-to-head points, excluding the first home game for the team \\
    & that has a greater number of home games \\
(6) & Goal differential, including shootout-deciding goals \\
(7) & Goals scored, including shootout-deciding goals \\
\hline
\end{tabular}
\caption{NHL tie-breakers. Each criterion is applied in ascending order until the tie is broken. Point percentage is the ratio of points earned to the maximum possible points. Goal differential is the goals scored minus the goals allowed. }
\label{tab:nhl-tiebreakers}
\end{table}

To determine if a team has mathematically clinched a playoff spot based on its current performance in the season (i.e., ``0-day lookahead'')
one must consider all possible outcomes of all remaining games. This makes it
a challenging combinatorial problem. In fact, it is known that this
problem, as well as other variants of the qualification problem, is
NP-Complete \cite{bernholt1999football, gusfield2002structure, kern2004computational}. Often, it is interesting to determine the outcomes of games (scenarios) over the next $n$ days that will result in a team provably clinching the playoffs (i.e., ``$n$-day lookahead''). These scenarios can be used by sports marketers to drive fan engagement and, in general, make following a team more exciting as the postseason approaches. We provide a simple example of 0-day and 1-day lookahead playoff clinching scenarios in \cref{tab:eastern-2024-04-15}.

\begin{table}[ht]
\centering
\begin{tabular}{l r r r r r r r r r l}
\hline
Team & PTS & MAX & GP & RW & OTW & SOW & DIFF & GF & GR & DIV \\
\hline
NYR & 114 & 114 & 82 & 43 & 8 & 4 & 53 & 282 & 0 & MET \\
CAR & 111 & 113 & 81 & 44 & 6 & 2 & 66 & 276 & 1 & MET \\
BOS & 109 & 111 & 81 & 36 & 7 & 4 & 45 & 266 & 1 & ATL \\
FLA & 108 & 110 & 81 & 41 & 7 & 3 & 65 & 263 & 1 & ATL \\
TOR & 102 & 106 & 80 & 33 & 8 & 5 & 45 & 297 & 2 & ATL \\
TBL & 96 & 98 & 81 & 36 & 5 & 3 & 21 & 285 & 1 & ATL \\
NYI & 92 & 94 & 81 & 28 & 9 & 1 & $-18$ & 241 & 1 & MET \\
WSH & 89 & 91 & 81 & 31 & 4 & 4 & $-38$ & 218 & 1 & MET \\
DET & 89 & 91 & 81 & 27 & 11 & 2 & 3 & 273 & 1 & ATL \\
PIT & 88 & 90 & 81 & 32 & 4 & 2 & 5 & 251 & 1 & MET \\
\hline
\end{tabular}
\caption{Eastern conference standings at the end of April 15, 2024. \\ \emph{0-day lookahead:} The New York Islanders (NYI) achieved a top-3 spot in the MET division and thus clinched the playoffs. In the worst-case scenario for NYI they will lose their remaining game and end with 92 points, which is more than the max points possible of all other MET teams except two. \\
\emph{1-day lookahead:} If WSH wins their final game in any fashion (2 points) they will take the final (wildcard) playoff berth from DET (the only ATL team contender), which could not overtake it even in its best-case scenario due to currently having four fewer regulation wins [tie-breaker (2)]. \\ 
\emph{Note: the standings of the six last teams were omitted for brevity.} \\ 
\emph{Columns: Current points (PTS), max points possible (MAX), games played (GP), regulation wins (RW), overtime wins (OTW), shootout wins (SOW), goal differential (DIFF), goals-for (GF), games remaining (GR), and division (DIV).}}
\label{tab:eastern-2024-04-15}
\end{table}

Determining playoff clinch scenarios for a team's games over the next $n$ days requires consideration of the combinations of outcomes in the relevant games during this time period. Naturally, the $n$-day lookahead problem is, in practice, algorithmically and computationally more challenging. In this paper we introduce an algorithm to address this challenge efficiently using a custom tree
search coupled with a constraint programming (CP)-based pruning method, preprocessing techniques, and node ordering heuristics. The methods presented here also lay the groundwork for analyzing
clinching scenarios for other achievements, including division titles or specific playoff seeds. 

The contributions of this paper are as follows:
\begin{itemize}
    \item We introduce a novel ``0-day lookahead'' approach, leveraging CP and decomposition, to determine whether a team has clinched the playoffs for some state of the regular season standings. Building on previous work, the approach accounts for the extensive set of tie-breaking rules used since the 2019-20 season. 
    \item We introduce a novel ``$n$-day lookahead'' algorithm for determining concrete playoff clinch scenarios, for a given team of interest, involving games to be played in the next $n$ days. The approach involves a custom tree search, bolstered by preprocessing techniques, pruning strategies, and node ordering heuristics.
    \item We conduct extensive benchmarking of our methods on the four NHL seasons from 2021-22 to 2024-25 and validate the accuracy of our ``$n$-day lookahead'' approach against scenarios (for $n \in \{0, 1\}$) previously published by the NHL, where we have an exact match. We report on algorithm runtime, pruning efficacy, and clinch scenario complexity. 
\end{itemize}

We start by reviewing the relevant literature in Section~\ref{sec:literature}. In Section~\ref{sec:zero_day_look_ahead} we introduce a ``0-day lookahead'' subroutine using constraint programming (CP) to determine if a team has clinched a playoff spot based on a current league state. In Section~\ref{sec:n_day_look_ahead} we introduce an ``$n$-day lookahead'' algorithm employing tree search to determine under which combinations of game outcomes in the next $n$ days a team will
clinch a playoff spot (if any), and present the results in Section~\ref{sec:results}. Finally, we present our conclusions in Section~\ref{sec:conclusion}.

\section{Relevant Literature}
\label{sec:literature}

The playoff clinch determination problem is common to many sports and has attracted significant attention in the academic community. A range of sports have been studied, including baseball
\cite{schwartz1966possible, robinson1991baseball, wayne2001new, adler2002baseball, kim2024improving}, soccer \cite{ribeiro2005application, lucena2008multi, raack2014standings}, car racing \cite{donne2012studying, raack2014standings}, (American) football \cite{whittle2014nfl}, basketball \cite{ito2018calculation, husted2019enhancing, husted2021improving}, and, as in this work, hockey \cite{russell2008mathematically,russell2009determining, russell2009lessons, russell2010computational, russell2012hybrid}. 

Each of these works solve different variations of the problem due to the variety of sports considered. Within a given sport, each league typically requires special consideration due to the variations in league structure, tie-breakers, point systems, types of game outcomes, and qualification criteria. Even when considering the same league, these factors might change over the years, as has happened multiple times in the NHL. There is also some variety in the exact question being answered: whether the team has clinched the playoffs already, the number of points or wins (also known as the ``magic number'' or the ``clinching number'') required to clinch the playoffs, or whether a specific ranking has been clinched.

The most relevant prior works to our own are those of Russell and van Beek who published a series of papers on clinching problems in the NHL for the 2005-06 and 2006-07 seasons \cite{russell2008mathematically,russell2009determining, russell2009lessons, russell2010computational, russell2012hybrid}. Although the league has changed considerably since then (30 teams vs. 32 now, 6 divisions vs. 4 now, top-8 in-conference qualification vs. top-3 in-division \& wild cards now, 3 tie-breakers vs. 7 now), these relevant works form a foundation for our methods. 

In Ref.~\cite{russell2008mathematically} they presented a ``0-day lookahead'' CP model for playoff clinching. In contrast to our 0-day approach (see Section \ref{sec:zero_day_look_ahead}), which iteratively solves a series of CP models and adds no-good constraints when necessary, they leverage customized propagation algorithms and dominance rules. In follow-up work~\cite{russell2009lessons, russell2009determining}, they presented an optimization model that determines the minimum number of points a team must get to clinch the playoffs, as well as the maximum number of points a team can get and still not qualify for the playoffs. Finally, in Ref.~\cite{russell2012hybrid} they extended their earlier works with an approach for points-needed-to-clinch (or almost clinch) that leverages CP, enumeration, network flows, and decomposition.

Our work is differentiated from these previous works in a number of key ways. First, our approach is tailored to modern league structure and significantly different tie-breakers (see \cref{tab:nhl-tiebreakers}), and introduces a novel goal assignment model to address this requirement. Second, following the NHL's own publication of ``lookahead'' scenarios for playoff clinching, our work builds on earlier efforts by determining a more detailed future result: the specific outcomes of games occurring in the next $n$ days that would cause a specific team to clinch the playoffs (if any). Finally, our approach is implemented in a modern solver and does not rely on special-purpose constraint propagation algorithms which were critical in Ref.~\cite{russell2008mathematically}. Instead, we leverage a relatively simple no-good constraint generation, recognizing that certain tie-breaking conditions are only very rarely triggered. 

\section{0-Day Lookahead Approach}
\label{sec:zero_day_look_ahead}

The 0-day lookahead approach determines if a team under consideration has clinched the playoffs for some state of the regular season standings.  It uses a CP model based on
Ref.~\cite{russell2008mathematically}, however, the approach has been updated
substantially to account for significant changes in the league structure,
qualification requirements, and tie-breakers.

The approach asks: ``can team $k$ be
eliminated?'' (from here on we refer to the team being tested as team $k$) and looks for an assignment of outcomes to the remaining games that results in elimination. If one is found, then the team has not clinched the
playoffs. Conversely, if one is not found, the team has clinched the
playoffs. An advantage of this formulation is that, in most scenarios, a team
being tested can be quickly shown not to clinch the playoffs by coming up with
a single counter-example. Furthermore, the constraint satisfaction problem is more convenient than the related optimization problem (e.g., ``find the worst rank that team $k$ could
get'').

For playoff clinching, it is sufficient to check the worst-case scenario for
team $k$. If there are no elimination scenarios in the worst-case scenario, then
there will not be any in any of the better scenarios. We use this assumption,
for example, by assuming that all teams playing against team $k$ win in regulation, giving them two points and team $k$ zero points.

\begin{figure}[!ht]
  \centering
  \captionsetup{margin=0cm}
  \includegraphics[width=0.75\textwidth]{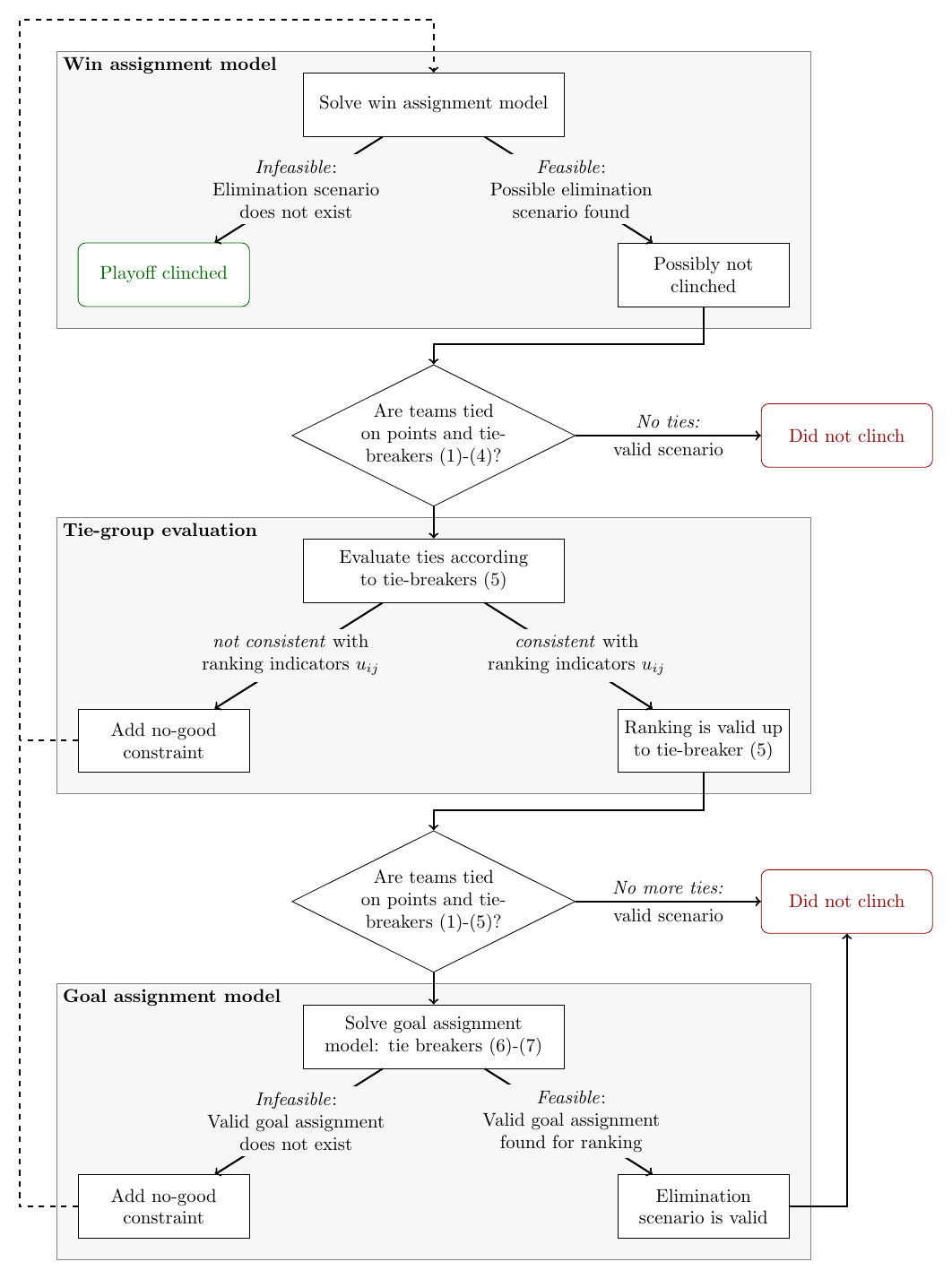}
  \caption{
    Summary of the 0-day lookahead clincher.
    Initially, the win assignment model is solved. If there are ties up to (and including) tie-breaker 
    (4), we check for consistency with tie-breaker (5) and solve the goal assignment
    model if necessary. Inconsistencies are remedied via no-good constraints and
    the revised win assignment model is re-run (forcing a different
    solution than previously seen). This continues until a valid elimination scenario is found or none are proven to exist.
  }
  \label{fig:zero_day_flow_chart}
\end{figure}

The full 0-day lookahead approach is presented in
Figure~\ref{fig:zero_day_flow_chart} and consists of three main modules: i) win assignment model, ii) tie-group evaluation, and iii) goal assignment model. In the vast majority of
cases, it is sufficient to solve only the win assignment model, described in
detail in Section~\ref{sec:win_assignment_model}. If the win assignment model is infeasible,
then an elimination scenario does not exist, and the team being considered has clinched
the playoffs. Alternatively, if an elimination scenario has been found and it
has no ties up to tie-breaker (4), then we know that the
playoffs have not been clinched (since we have a counter-example in hand).

Ties complicate matters due to the elaborate tie-breakers in the
NHL. Specifically, tie-breaker (5) is different and more problematic than the others. Rather than being applied to pairs of tied teams, it is applied to groups of tied teams of arbitrary size. Modeling this tie-breaker directly could be done by enumerating all possible tie groups and adding appropriate constraints. However, given the large number of possibilities, this would have significantly slowed down the model. Instead, we add some additional variables representing a ``guess'' at the ranking of teams tied beyond the tie-breaker (4), and then use lazy constraint generation to generate no-good constraints if the guess was incorrect in a procedure resembling logic-based decomposition~\cite{hooker2011logic}. 

For example, assume that teams A, B, and C are tied on points and tie-breakers (1)--(4), and that the solver assigns a ranking guess $u_{AB}=1, u_{AC}=1, u_{BC}=0$ (i.e., $A > B$, $A > C$, and $C > B$). The algorithm then evaluates the head-to-head points among {A, B, C} according to tie-breaker (5). If the actual head-to-head ranking yields a different ranking, for example A > C > B, then the guess is inconsistent. A no-good constraint excluding this specific assignment is added, and the model is re-solved.

These ``guess'' variables are initially unconstrained, allowing the solver to assign any ranking among teams tied on tie-breakers (1)--(4). When a solution is found, the algorithm conducts a tie-group evaluation (Figure~\ref{fig:zero_day_flow_chart}) and checks for consistency between the effective ranking (resolving all tie-breakers) and the solver-assigned variable values. An
inconsistency indicates that tie-breakers (5)--(7), which were not previously considered,
are guaranteed to result in a different ranking than one presumed by the solver
-- either by ``direct points'' (5) or by reasoning about potential goals
(6)--(7) in the  separate goal assignment model (detailed in
Section~\ref{sec:goal_assignment_model}). In this case, we add a ``no-good
constraint'' to exclude this specific assignment of wins and ranking values, forcing the solver to explore an alternative solution in the next run. This process continues until a consistent ranking is found or the model becomes infeasible.

\subsection{Win Assignment Model}
\label{sec:win_assignment_model}

The following section details the win assignment model. This model explores presumed rankings based on
assigned game \emph{outcomes} (i.e., one of RW, OTW, SOW, SOL, OTL, RL for each
remaining game), while not assigning goals to the games yet.
\bigskip

\noindent \textbf{Parameters.} Let $T$ be the set of NHL teams, $C_i$ represent the set of teams in the same conference as team $i$ (including team $i$), and $D_i$ the set of teams in the same division as team $i$ (including team $i$). We let $k$ represent the index of the team under evaluation for playoff clinching.  We let parameters $p^0_i, p^0_{ij}, w^0_i, w^0_{ij}$ represent the current points, points earned against team $j \neq i$, total wins, and wins over team $j \neq i$ for team $i \in T$, respectively. Further, we let $w^{R,0}_i, w^{O,0}_i, w^{S,0}_i$ be the current regulation, overtime, and shootout wins for team $i \in T$, respectively. $w^{R,0}_{ij}, w^{O,0}_{ij}, w^{S,0}_{ij}$ represent the current regulation, overtime, and shootout wins over team $j \neq i$ for team $i$. Finally, $g_i$ is the number of remaining games for team $i$ and $g_{ij}$ the number of games remaining between teams $i$ and $j$.

\bigskip
\noindent \textbf{Variables.} The primary decision variables in our win assignment model are integer-valued game outcome variables $x^R_{ij}, x^O_{ij}, x^S_{ij} \in \{0, \dots, g_{ij}\}$ that indicate the number of regulation, overtime, and shootout wins by team $i$ over team $j$ in their remaining games. The model also has binary variables tracking the ranking and performance of teams. $b_{ij} \in \{0,1\}$ is 1 if team $i$ outranks team $j$, and 0 otherwise. As previously discussed, $u_{ij} \in \{0,1\}$ represents the model's ``guess'' as to whether $i$ outranks $j$ for tie-breakers (5) and beyond (if relevant). $t_i \in \{0,1\}$ is 1 if team $i$ is a top-3 team in its division and 0 otherwise, and $d_i \in \{0, 1\}$ is 1 if team $i$ is a top-3 team that is also outranked by team $k$, and 0 otherwise.\footnote{The definition of $d_i$ is implicitly always with respect to the team
being considered for evaluation (team $k$), therefore it does not carry a
separate subscript for the ``other'' team.} 

To ease modeling, we introduce expressions over the main decision variables. These include $p_{ij}$, the total points earned against team $j \neq i$ by team $i$ at the end of the season, and $p_i$ which represents the point total for team $i$ at the end of the season, as follows:
\begin{equation}
  p_{ij} = p^0_{ij} + 2x^R_{ij} + 2x^O_{ij} + 2x^S_{ij} + x^O_{ji} + x^S_{ji}, \quad p_i = \sum_{j \neq i} p_{ij}
\end{equation}
We also define expressions $w_i, w^R_i, w^O_i, w^S_i$, representing total end-of-season wins, regulation wins, overtime wins, and shootout wins, respectively:
\begin{equation}
  w_i = w^R_{i} + w^O_{i} + w^S_{i}, \quad   w^R_{i} = \sum_{j \neq i} w^{R,0}_{ij} + x^R_{ij}, \quad 
  w^O_{i} = \sum_{j \neq i} w^{O,0}_{ij} + x^O_{ij}, \quad 
  w^S_{i} = \sum_{j \neq i} w^{S,0}_{ij} + x^S_{ij}
\end{equation}
Finally, we define expressions $w^R_{ij}, w^O_{ij}, w^S_{ij}$ to indicate the total regulation, overtime, and shootout wins that team $i$ achieved over team $j$: 
\begin{equation}
  w_{ij} = w^{R,0}_{ij} + w^{O,0}_{ij} + w^{S,0}_{ij} + x^R_{ij} + x^O_{ij} + x^S_{ij}
\end{equation}

\noindent \textbf{Constraints.} The first family of constraints in our model are the \emph{worst-case assumption} constraints. Namely, we express that team $k$ loses all of its remaining games in regulation:\footnote{It can be readily seen that a regulation loss is the worst possible outcome for team $k$, because it awards the least number of points to team $k$.}
\begin{equation} 
\label{eq:worst-case}
\forall \, i \in T, i \neq k: \quad x^R_{ki} = x^O_{ki} = x^S_{ki} = 0, \quad x^R_{ik} = g_{ik},  \quad x^O_{ik} = x^S_{ik} = 0, \quad t_k = 0
\end{equation}
The last constraint ($t_k = 0$) expresses that team $k$ cannot be a top-3 team,  otherwise it would qualify directly (and we are looking for an elimination scenario).

Next, all remaining games must have an outcome (i.e., won either by team $i$ or team $j$):
    \begin{equation}
      \label{eq:has-outcome}
      \forall \, i,j \in T, i\neq j : \quad 
        x^R_{ij} + x^O_{ij} + x^S_{ij}
        + x^R_{ji} + x^O_{ji} + x^S_{ji}
        = g_{ij}\,
    \end{equation}
    This requirement applies to all games, whether they involve team $k$ or not.
    If one of the teams is $k$ (e.g., $j=k$), then the assumptions in Eq.~\eqref{eq:worst-case}
    reduce Eq.~\eqref{eq:has-outcome} to the assumed regulation losses $x^R_{ik} = g_{ik}$.

Next, we define ranking variable $b_{ij}$. The ranking is symmetrical, so we only define $b_{ij}$ for $i<j$. After this definition, when we refer to $b_{ij}$, we mean it literally if $i<j$,
otherwise (i.e., if $i>j$) we implicitly mean $1-b_{ij}$ (this simplifies the
presentation of the equations). The total points criterion and tie-breakers
(2)--(4) then imply the following relation between points/wins and the
ranking indicators:
\begin{align}
  \forall \, i,j, i < j : \quad
  &(p_i > p_j) \\
  &\lor [(p_i = p_j) \land (w^R_{i} > w^R_{j})] \notag \\
  &\lor [(p_i = p_j) \land (w^R_{i} = w^R_{j}) \land (w^O_{i} > w^O_{j})] \notag \\
  &\lor [(p_i = p_j) \land (w^R_{i} = w^R_{j}) \land (w^O_{i} = w^O_{j}) \land (w^S_{i} > w^S_{j})] \notag \\
  &\lor [(p_i = p_j) \land (w^R_{i} = w^R_{j}) \land (w^O_{i} = w^O_{j}) \land (w^S_{i} = w^S_{j}) \land u_{ij}] \notag \\ 
  &\iff b_{ij} = 1\, \notag
\end{align}
The last term involves $u_{ij}$, and represents the situation when the total points and all win counts are equal [tied up to tie-breaker
(4)], and the model's ``guess'' for $u_{ij}$ determines the ranking according to further tie-breakers (validated in the next step of the algorithm flow). 

Next, we include constraints relating to variable $t_i$. Team $i$ is top-3 iff it outranks 5 or more teams in its division:
\begin{equation}
  \forall \, i \in T : \quad t_i = 1 \iff \sum_{j \in D_i, j \neq i} b_{ij} \geq 5\,
\end{equation}
There are exactly three top-3 teams in each division $D$:
\begin{equation}
  \forall \, D : \quad \sum_{i \in D} t_i = 3\,.
\end{equation}
And we ensure that non-top-3 teams do not outrank any of the top-3 teams:
\begin{equation}
  \forall i,j \in D_i, i\neq j : \quad b_{ji} \Rightarrow (t_i=0) \lor (t_j=1)\,
\end{equation}
For example, if $i$ and $j$ are in the same division and $j$ outranks $i$ then it cannot be that $i$ is in the top-3 ($t_i=1$) and $j$ is simultaneously \emph{not} in the top-3 ($t_j=0$).

Next, we include constraints relating variable $d_i$ to variable $t_i$: 
\begin{equation}
\forall i \in C_k, i\neq k: \quad d_i=1 \iff (t_i=1) \land (b_{ki}=1)\,
\end{equation}
 i.e., $d_i$ is 1 exactly when it is in the top-3 of its own division (indicated
by $t_i=1$) \textit{and} team $k$ is simultaneously ahead of it ($b_{ki}=1$). Note that
team $i$ and $k$ are not necessarily in the same division here; this is leveraged to correctly capture situations where teams ranked lower than $k$ in the conference still qualify because they are top-3 in their division in the next step.

Finally, we enforce that team $k$ does not qualify via a wildcard ticket:
\begin{equation}
  \label{eq:no-wildcard}
  \sum_{i \in C_k} (b_{ik} + d_i) \geq 8\,,
\end{equation}
where $d_i=1$ if team $i$ is in the top-3 of the other division \emph{and} is outranked by team $k$. For example, if team $k$ outranks all teams in the other division (so $\sum_i d_i=3$), then in order to be eliminated, it must be outranked by at least five teams in its division ($\sum_{i}b_{ik} \geq 5$) -- the teams ranked 4th and 5th in its division would take the two wild card spots. 

\subsection{Goal Assignment Model}
\label{sec:goal_assignment_model}

The purpose of this model is to find a valid goal assignment for the game outcomes and ranking returned by the win assignment model that has, at this point, been verified for tie-breakers (1) through (5). This model is only constructed and solved when there is a relatively rare tie up to (and including) the 5th tie-breaker in a prospective elimination scenario found by the win assignment model (see Figure~\ref{fig:zero_day_flow_chart}). We detail the parameters, variables and constraints of this model here.

\bigskip 
\noindent \textbf{Parameters.} Let $T^G$ be the set of teams tied on points and first tie-breakers (1) through (5) (``tie group'').  Let $P$ be the set of pairs $(i, j) \in T \times T$ such that: $i \neq j$, at least one team is in $T^G$, and $g_{ij}\geq 1$. Let $\delta_i^0, f_i^0$ be the goal differential and total goals scored, respectively, for team $i \in T^G$ for games involving no tied teams (the goals in games involving at least one tied team are solved for by this model). Finally, let $\hat{x}_{ij}^R, \hat{x}_{ij}^O, \hat{x}_{ij}^S$ be the number of regulation, overtime, and shootout wins by team $i$ over team $j$, as obtained in the feasible solution returned from the win assignment model, and $\hat{u}_{ij}$ the ranking returned.

\bigskip
\noindent \textbf{Variables.} The model includes decision variables $s_{ij\ell}$ representing the score of team $i$ vs. team $j$ in game $\ell$, where $(i,j) \in P$, $\ell \in [1,g_{ij}]$, $s_{ij\ell} \in [0,99]$, where 99 is an upper bound on the number of goals chosen to be large enough such that exceeding it would be extremely unlikely. Variables $y_{ij\ell}^R, y_{ij\ell}^O, y_{ij\ell}^S$ are Boolean indicators for regulation, overtime, and shootout wins by team $i$ over team $j$ in game $\ell$. 

We also introduce expressions over the decision variables to ease modeling. Specifically, goal differential for tied team $i \in T^G$ at the end of the season ($\delta_i$) and total goals scored by tied team $i \in T^G$ at the end of the season ($f_i$):
\begin{equation}
  \forall \, i \in T^G: \quad \delta_i = \delta_i^0 + \sum_{j \in T} \sum_{\ell \in [1,g_{ij}]} (s_{ij\ell} - s_{ji\ell}), \quad f_i = f_i^0 + \sum_{j \in T} \sum_{\ell \in [1,g_{ij}]} s_{ij\ell}
\end{equation}

\noindent\textbf{Constraints.} The first set of constraints ensure that each game must end with one team winning in one of three possible ways:
\begin{equation}
  \forall \, (i, j) \in P, i < j, \ell \in [1, g_{ij}]: \quad y_{ij\ell}^R +
    y_{ij\ell}^O + y_{ij\ell}^S + y_{ji\ell}^R + y_{ji\ell}^O + y_{ji\ell}^S = 1
\end{equation}

Next, we link the $y$ variables to the outcomes from the win assignment model:
\begin{equation}
      \forall \, (i, j) \in P: \quad
      \sum_{\ell \in [1,g_{ij}]} y_{ij\ell}^R = \hat{x}_{ij}^R, \quad 
      \sum_{\ell \in [1,g_{ij}]} y_{ij\ell}^O = \hat{x}_{ij}^O, \quad 
      \sum_{\ell \in [1,g_{ij}]} y_{ij\ell}^S = \hat{x}_{ij}^S 
    \end{equation}

Then we ensure that scores must match the win type: regulation wins have unrestricted goal differential while overtime and shootout wins require a one-goal margin:
    \begin{equation}
      \forall \, (i, j) \in P, \ell \in [1, g_{ij}]: \
      y_{ij\ell}^R \Rightarrow s_{ij\ell} > s_{ji\ell}, \ 
      y_{ij\ell}^O \Rightarrow s_{ij\ell} - s_{ji\ell} = 1, \ 
      y_{ij\ell}^S \Rightarrow s_{ij\ell} - s_{ji\ell} = 1 
    \end{equation}

Next, we impose that the ranking of teams tied on tie-breakers (1)--(5), based on goal differential and total goals scored, should match those from the win assignment model ($\hat{u}_{ij}$):
    \begin{equation}
      \forall \, i, j \in T^G, i < j : \quad (\delta_i > \delta_j) \vee [(\delta_i = \delta_j)
        \wedge (f_i > f_j)] \iff \hat{u}_{ij} = 1
    \end{equation}
    
Finally, a ``total tie'', in which two or more teams are tied on points and all tie-breakers is theoretically possible. However, the odds are extremely low of this happening, and if it does, the rules do not specify how to break the tie. Thus, we add a constraint to prevent teams tied on tie-breakers (1)--(5) from also being tied on both goal differential and total goals:
\begin{equation}
  \forall \, i, j \in T^G, i < j : \quad (\delta_i = \delta_j) \Rightarrow (f_i \neq f_j)
\end{equation}

\noindent\textbf{Objective.}  An objective is not strictly required. However, without it, the solver could return goal assignments with very high and therefore unrealistic scores.
Therefore, we add an objective -- minimizing the sum of goals:
\begin{equation}
  \min \sum_{(i,j) \in P} \sum_{\ell \in [1,g_{ij}]} s_{ij\ell}
\end{equation}

\section{\texorpdfstring{$n$}{n}-day Lookahead Approach}
\label{sec:n_day_look_ahead}

This section describes a tree search algorithm for determining under which
outcomes of the games in the next $n$ days an NHL team will clinch the playoffs. A simplified pseudo code for this algorithm is presented in \cref{alg:n_day_clinching}. As an illustrative example, consider the 1-day clinching scenario for the Montreal Canadiens on April 15, 2025 (\cref{fig:example_scenarios}). The tree has a single relevant game (Columbus vs. Philadelphia). Five of the six outcomes for this game lead to team $k$ (Montreal) clinching the playoffs -- confirmed by the 0-day clincher. The algorithm prunes sideways, yielding the single-literal clinching scenario. 

The games and outcomes are organized in a hierarchical tree structure where
each node represents an outcome of a particular game (see
Figure~\ref{fig:pruning_right}). The outcomes are arranged from lowest value
(regulation loss -- RL) on the left, to highest value (regulation win -- RW) on
the right, relative to team $k$.\footnote{In overtime, if a team pulls its goalie and then concedes a goal, the team does not receive a point for their loss. In extremely rare cases these additional outcomes are relevant to the search, and could be included via additional game outcome nodes. However, due to the rarity of this occurrence, we leave this to future work.} Following any path from the root [see CreateRootNode()] to a node
provides a complete description of the assumed outcomes for all games along
that path, allowing for systematic exploration of all possible game result
combinations. 

After a presolve phase (see \cref{sec:presolve_phase}), the algorithm searches the full space of possible outcomes by processing the nodes in a queue. For each node, the games are preprocessed (see \cref{sec:game_preprocessing}), and then a set of inference algorithms attempt to prune the tree [see CanPruneWith()]. If any of them succeed, the tree is pruned below that node and sideways if possible (see \cref{sec:tree_pruning}). If all fail, then the children of that node are added to the queue [see GetChildren()]. The tree search is controlled by game, node, and inference algorithm ordering functions (see \cref{sec:ordering_functions}). At the final step, equivalent clinching scenarios are consolidated [see ConsolidateClinchingScenarios()]. 

An alternative approach would be to try and solve the $n$-day lookahead problem using a single model build-and-solve. The presented 0-day model answers ``has team $k$ clinched the playoffs?'' by finding the existence of an elimination scenario. We could ask ``has team $k$ been eliminated from the playoffs?'' by trying to find the existence of a clinch scenario. Then, we could adjust the model to output all feasible clinch scenarios (as well as modify the model variables to track explicit game outcomes versus aggregated points), which is often considerable, and down-select to the scenarios that involve a clinch in the next $n$-days. The primary issue here is that this approach would generate many more scenarios than we actually care about, and, in all likelihood, take much more time. Our proposed approach gives us a high level of control over the specific outcomes that are explored, and leverages existing literature that provides a foundation for our work. 

\begin{algorithm}[t!]
\footnotesize
\caption{$n$-day Lookahead Clinching Algorithm}
\label{alg:n_day_clinching}

\KwIn{Team $k$, list of games in next $n$ days}
\KwOut{Set of clinching paths (game outcomes leading to clinch)}

\tcp{Presolve phase}
$games \gets$ PreprocessGames($games$, team $k$, $league$) \\
\If{\textbf{not} \textnormal{ClinchesPlayoffsInBestCase}(\textnormal{team} k, games)}{
    \Return{$\emptyset$}
}
$inferenceAlgorithms \gets$ [HasClinchedPlayoffs, ClinchesPlayoffsInBestCase] \\
\If{\textnormal{ClinchesEliminationInWorstCase}(\textnormal{team} k, games)}{
    $inferenceAlgorithms$ += [HasClinchedElimination]
}
\tcp{Initialize tree search}
$clinchingNodes \gets \emptyset$ \\
$root \gets$ CreateRootNode(team $k$) \\
$root.game \gets$ GameOrderingFunction($games$, $root$, team $k$) \\

$queue \gets$ GetChildren($root$, team $k$) \\

\tcp{Main loop over nodes}
\While{$queue \neq \emptyset$}{
    $node \gets$ NodeOrderingFunction($queue$, team $k$) \\
    $queue \mathbin{{-}{=}} node$ \\
    
    $expandChildren \gets$ \textbf{true} \\
    $league \gets$ GetLeagueStateAtNode($node$) \\
    $inferenceOrderForNode \gets$ InferenceAlgorithmOrderingFunction(\\
        \hspace{1cm} $inferenceAlgorithms$, $queue$, $node$, team $k$, $league$)

    \tcp{Use inference algorithms to prune below and sideways}  
    \ForEach{$algorithm \in inferenceOrderForNode$}{
        \If{\textnormal{CanPruneWith}(algorithm, queue, node, \textnormal{team} k, league)}{
            \If{algorithm = \textnormal{HasClinchedPlayoffs}}{$clinchingNodes$ += $node$}

            \If{node.game.orderedOutcomes}{
            $direction \gets$ \textnormal{GetDirection}($algorithm$)  \\
            $queue \gets$ \textnormal{PruneSideways}($queue$, $node$, $direction$) }
            $expandChildren \gets$ \textbf{false} \\
            \textbf{break}
        }   
    }
    \tcp{Expand node and add children to queue}
    \If{expandChildren \textbf{and not} \textnormal{IsLeafNode}($node$)}{
        $children \gets$ GetChildren($node$, team $k$) \\
        \ForEach{$child \in children$}{
        $childLeague \gets$ GetLeagueStateAtNode($child$) \\
        $games \gets $ PreprocessGames($games$, team $k$, $childLeague$) \\
        $child.game \gets$ GameOrderingFunction($games$, $child$, team $k$)
        }   
        $queue$ += $children$}
}
\tcp{Return consolidated clinching scenarios}
\Return{\textnormal{ConsolidateClinchingScenarios}(clinchingNodes)}
\end{algorithm}

\subsection{Presolve Phase}
\label{sec:presolve_phase}

Before initiating the tree search, two preliminary checks are performed:

\begin{enumerate}

\item \textbf{Best-case Check:} Use the 0-day playoff clincher to evaluate an impossibly-optimistic scenario [see ClinchesPlayoffsInBestCase()]
where team $k$ wins all games in the next $n$ days and all other teams lose their games simultaneously (impossible in practice). If team $k$ does not clinch the
playoffs in this scenario, it cannot clinch in any realistic scenario, making
the tree search unnecessary.

\item \textbf{Worst-case Check:} Use the 0-day elimination clincher\footnote{The 0-day elimination clincher is similar to the 0-day playoff clincher, but it looks for a feasible qualifying scenario in team $k$'s best-case, instead of a feasible elimination scenario in team $k$'s worst-case.} to evaluate an impossibly-pessimistic scenario [see ClinchesEliminationInWorstCase()]
where team $k$ loses all games in the next $n$ days and all other teams win their
games simultaneously (impossible in practice). If team $k$ does not clinch
elimination in this scenario, it cannot clinch elimination in any realistic
scenario, making the elimination clinching inference algorithm unnecessary
during the tree search.

\end{enumerate}

\subsection{Game Preprocessing}
\label{sec:game_preprocessing}

To minimize the search space and search it efficiently, games are preprocessed
at each node [see PreprocessGames()]. This is a key detail that allows us to use all the available
information on outcomes of the games in question to either prune games if
they are irrelevant, or at least to order their outcomes (from the point of
view of team $k$) which could later allow us to prune sideways (see \cref{sec:tree_pruning}).

Before detailing the rules for preprocessing, we state the following: a team
$i$ (that is not team $k$) is marked as ``out of the race'' with team $k$ if any
of these conditions are met: it is in the other conference, it has clinched
the playoffs or elimination, it outranks team $k$ (see definition below), or it
is outranked by team $k$. If in team $i$'s worst-case scenario (lose all
remaining games in regulation) it still has a higher rank than team $j$ with
the best-case scenario (win all remaining games in regulation), then we say
that team $i$ has outranked team $j$. Note that as the season progresses and teams clinch the playoffs or elimination, they are marked as ``out of the race''.

With those definitions in place, we can now discuss game preprocessing. The
best case is if both teams are out of the race; the game is then excluded as
irrelevant. When one team in a game is out of the race (with respect to team $k$), the outcome ordering from $k$'s perspective is unambiguous. The ordering is based on points and an overtime win being a stronger outcome than a shootout win (see tie-breakers). The outcomes for team $k$ can be seen in \cref{fig:pruning_right} where they are ordered from weakest (RL, on the left) to strongest (RW, on the right). This monotonicity allows sideways pruning, since if $k$ clinches under a given outcome, it will also clinch under any stronger outcome. If instead both teams are in the race, an outcome that is stronger for one may be better or worse for team $k$, making the ordering from $k$'s perspective ambiguous, so we cannot prune sideways. 

\subsection{Tree Pruning}
\label{sec:tree_pruning}
The tree search employs pruning strategies based on inference algorithms
(different runs of the 0-day clincher) to efficiently explore the solution
space. If team $k$ clinches the playoffs at a given node [see HasClinchedPlayoffs()], we always prune the nodes
below that node, since any additional game outcomes will necessarily also
result in clinching the playoffs. Additionally, if the game outcomes have a
well-defined ordering (see below), we can also prune right (see
Figure~\ref{fig:pruning_right}) because higher-value outcomes will also lead
to clinching the playoffs [see PruneSideways()]. Conversely, let us assume that we are at a node where team $k$ will not clinch
the playoffs, based on clinching elimination at that node [see HasClinchedElimination()], or based on not
clinching the playoffs even in the impossibly optimistic best-case [see ClinchesPlayoffsInBestCase()]. In such a case we can definitely prune below,
as well as left if the game outcomes have a well-defined ordering. Games involving two teams that are not out of the race do not have a well-defined ordering of outcomes from the point of view of team $k$, which means we cannot prune sideways.

\begin{figure}[t]
  \centering
  \includegraphics[width=0.7\textwidth]{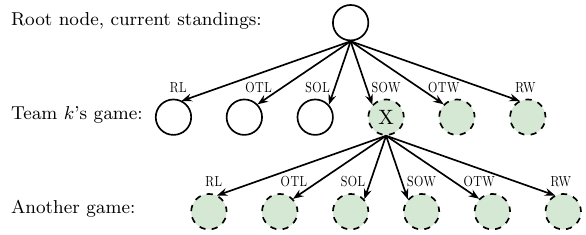}
  \caption{
    An example of pruning right. In this example, assume that team $k$ clinches
    the playoffs (such nodes have dashed borders and are shown in green) if it gets an SOW outcome in
    the first game. Then we can prune below -- since additional outcomes will not change this fact, as
    well as right -- since any stronger result would result in the same conclusion.
  }
  \label{fig:pruning_right}
\end{figure}

\subsection{Ordering Functions}
\label{sec:ordering_functions}

The tree search is controlled by three types of ordering functions. Below we
provide an explanation of each and the default implementation we have used;
additional benchmarking and experimentation might lead to better choices for
these components.

\begin{enumerate}
  \item \textbf{Game Ordering} --
    Determines which game to branch on next at a non-pruned node [see GameOrderingFunction()]. Our algorithm
    promotes games involving teams closest to being outranked by team $k$ if we
    have not yet pruned nodes belonging to the remaining games, or otherwise
    promotes games for which nodes have already been pruned (in the current
    run).

  \item \textbf{Inference Algorithm Ordering} --
    Determines the sequence of inference algorithms to try at each explored
    node [see InferenceAlgorithmOrderingFunction()]. If an inference algorithm succeeds in pruning, the remaining
    inference algorithms are skipped for that node. Similar to the above, the
    inference algorithms are ordered according to the number of times we have
    been able to use them to prune the tree (in the current run).

  \item \textbf{Node Ordering} --
    Determines the next node to explore in the tree search [see NodeOrderingFunction()]. Our algorithm uses
    a form of depth-first search motivated by wanting to successfully prune the
    tree early on, in order to improve the other ordering methods' odds of
    pruning (since they are based on dynamic ordering).
\end{enumerate}

\section{Results}
\label{sec:results}

In this section we present the results of $n$-day NHL playoff clinching determination experiments conducted on the seasons 2021-22 through 2024-25 using data pulled from the NHL API \cite{nhl_api}, with $n \in \{0,1,2,3\}$. All experiments were run on a MacBook Pro, 2.6 GHz 6-Core Intel Core i7, with 16 GB RAM, using Python 3.10.2 and CP-SAT from OR-Tools 9.11.4210 \cite{cpsat}. CP-SAT was chosen due to its strong performance, as well as the expressivity provided by \texttt{OnlyEnforceIf()}, which was used to express logical relationships between integer comparisons and Boolean indicators for ranking, tie-breaking, and qualification constraints. The start dates for each season were chosen just before the first clinch for back-testing convenience. All clinching scenarios for $n \in \{0, 1\}$ were validated successfully by comparing to the official scenarios published on the NHL website (for example, see Ref.~\cite{nhl2025clinching}). 

\subsection{0-Day Lookahead Results}

\begin{figure}[t]
  \centering
  \begin{subfigure}[b]{0.49\textwidth}
    \centering
    \caption{}
    \label{fig:zero_day_clinches}
    \includegraphics[width=\textwidth]{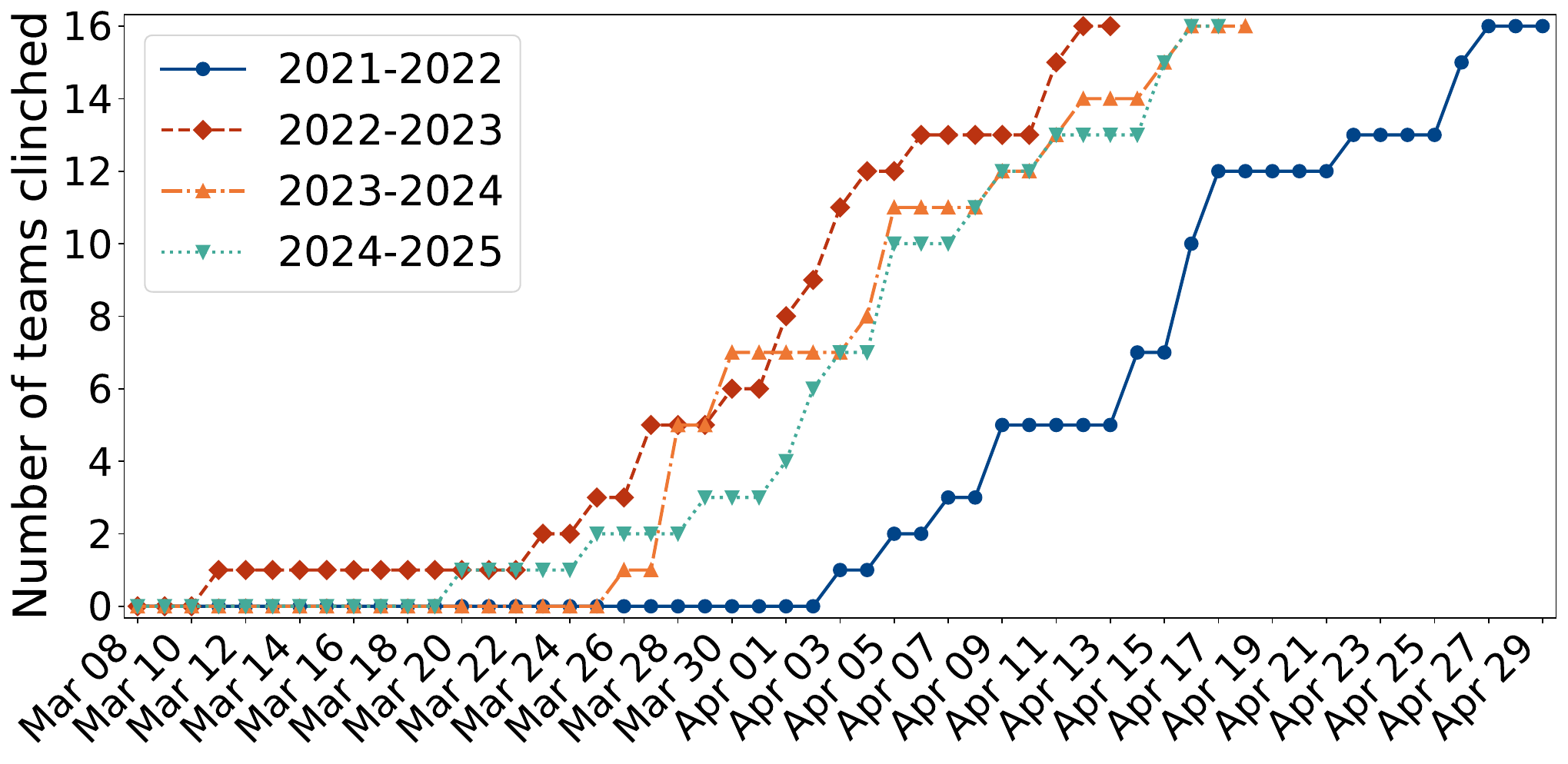}
  \end{subfigure}
  \begin{subfigure}[b]{0.49\textwidth}
    \centering
    \caption{}
    \label{fig:zero_day_solve_times}
    \includegraphics[width=\textwidth]{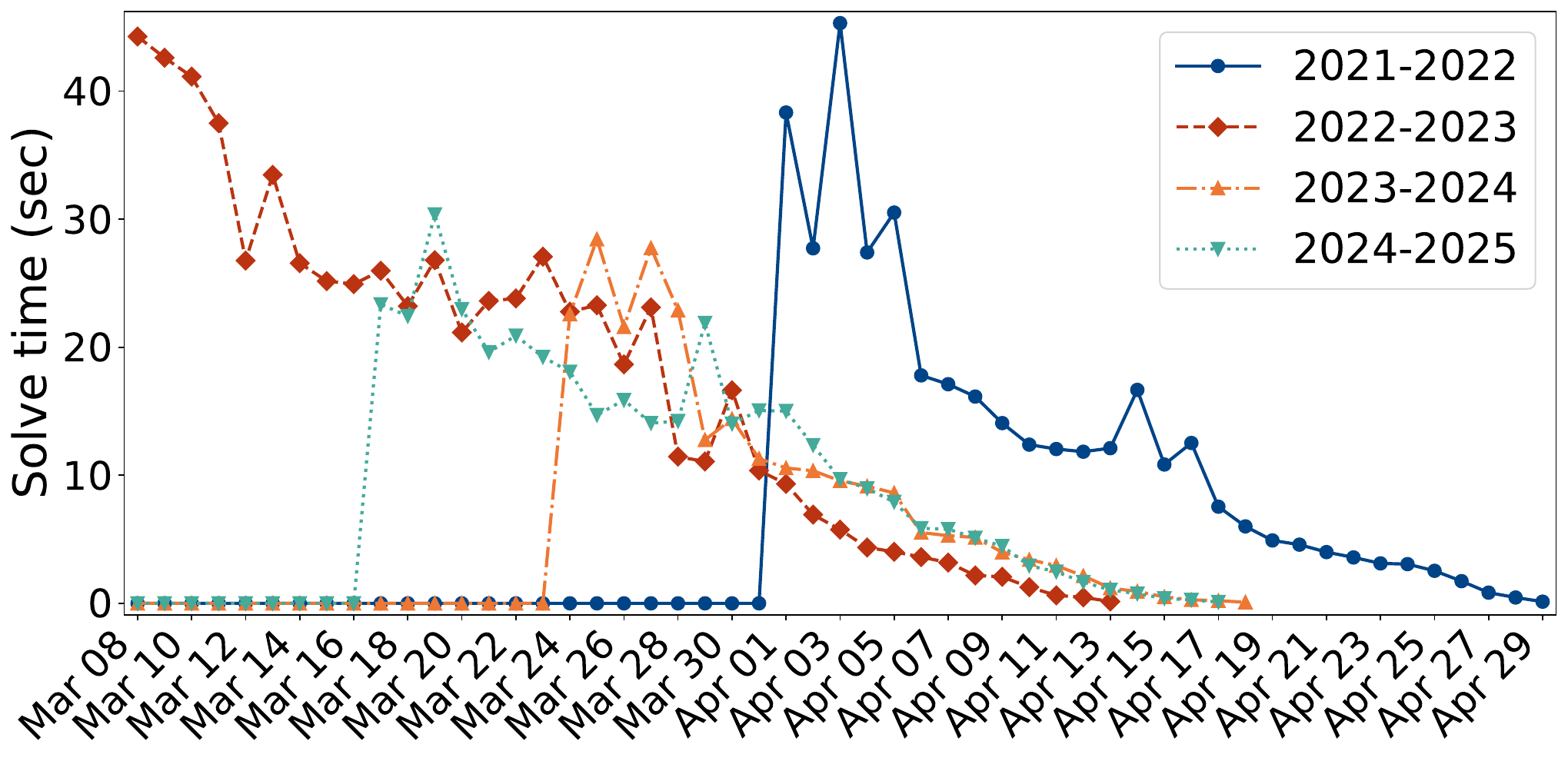}
  \end{subfigure}
  \caption{0-day lookahead approach experiments. \textbf{(a)} Number of teams that clinched the NHL playoffs at each date, for four seasons. \textbf{(b)} Total solve time required by our implementation to determine the playoff clinches (or a no clinch result) at each date.}
  \label{fig:zero_day}
\end{figure}

\cref{fig:zero_day_clinches} shows the accumulation of 0-day playoff clinches, which follow a similar pattern over each of the four NHL seasons. The schedule of the 2021-2022 season was shifted later due to the COVID-19 pandemic, explaining the right-shift evident in the figure. In the 2022-2023 season, the Boston Bruins clinched the playoffs early, accounting for the earlier bump seen on the left. \cref{fig:zero_day_solve_times} shows the total solve time for each date. We expect to find that the difficulty, and hence solve time, are low early (when few teams can clinch) and late in the season (when most playoff spots have been filled), and higher in the middle, when teams are on the verge of clinching. This is generally observed, with the early first clinch in 2022-2023 presumably leading to the early peak seen that year. 

\subsection{1-Day Lookahead Results}

1-day lookahead clinching scenarios for a specific team can be described in Disjunctive Normal Form (DNF), in which each literal is a game with a set of included outcomes. The DNF format naturally emerges from our tree search, since each root-to-leaf clinching path becomes a conjunction, and the disjunction of all such paths forms the scenario. This format also matches the style used by the NHL in their official publications \cite{nhl2025clinching}.

The complexity of the clinching scenarios can be quantified by counting the total number of literals. For example, in \cref{fig:example_scenarios} we present the 1-day playoff clinching scenarios produced for April 15th, 2025. These scenarios are the direct output from our implementation, and match the format used by the NHL when publishing official clinching scenarios. They are representative in that they provide examples of clinching scenarios with varying levels of complexity, and match with the NHL's official clinching scenarios for that day \cite{nhl2025clinching}. The total literal count for each of the three clinching scenarios shown is 1, 3, and 8, respectively. 

\begin{figure}[t]
\footnotesize
\centering
\begin{tcolorbox}[
  colback=gray!8,
  colframe=gray!50,
  arc=2pt,
  boxrule=0.5pt,
  left=8pt, right=8pt, top=6pt, bottom=6pt,
  width=\linewidth
]

\textbf{Eastern Conference}

\textit{Montreal Canadiens} will clinch a playoff berth if:
\begin{itemize}
  \item The Columbus Blue Jackets get any result except a regulation win
    against the Philadelphia Flyers.
\end{itemize}

\smallskip
\textbf{Western Conference}

\textit{Minnesota Wild} will clinch a playoff berth if any of the following holds:
\begin{itemize}
  \item They get at least one point against the Anaheim Ducks; \textbf{or}
  \item The St.\ Louis Blues lose to the Utah Hockey Club in any fashion; \textbf{or}
  \item The Calgary Flames lose to the Vegas Golden Knights in any fashion.
\end{itemize}

\textit{St. Louis Blues} will clinch a playoff berth if any of the following holds:
\begin{itemize}
  \item They defeat the Utah Hockey Club in regulation; \textbf{or}
  \item They defeat the Utah Hockey Club in any fashion \textbf{and} the
    Calgary Flames get any result except a regulation win against the Vegas
    Golden Knights; \textbf{or}
  \item They get at least one point against the Utah Hockey Club \textbf{and}
    the Calgary Flames lose to the Vegas Golden Knights in any fashion; \textbf{or}
  \item They defeat the Utah Hockey Club in any fashion \textbf{and} the
    Minnesota Wild lose to the Anaheim Ducks in regulation; \textbf{or}
  \item The Calgary Flames lose to the Vegas Golden Knights in regulation.
\end{itemize}

\end{tcolorbox}
\vspace{-2mm}
\caption{Generated 1-day lookahead playoff clinching scenarios for April 15th, 2025.}
\label{fig:example_scenarios}
\end{figure}

\begin{figure}[h!]
  \centering
  \begin{subfigure}[b]{0.49\textwidth}
    \centering
    \caption{}
    \label{fig:one_day_elapsed_time}
    \includegraphics[width=\textwidth]{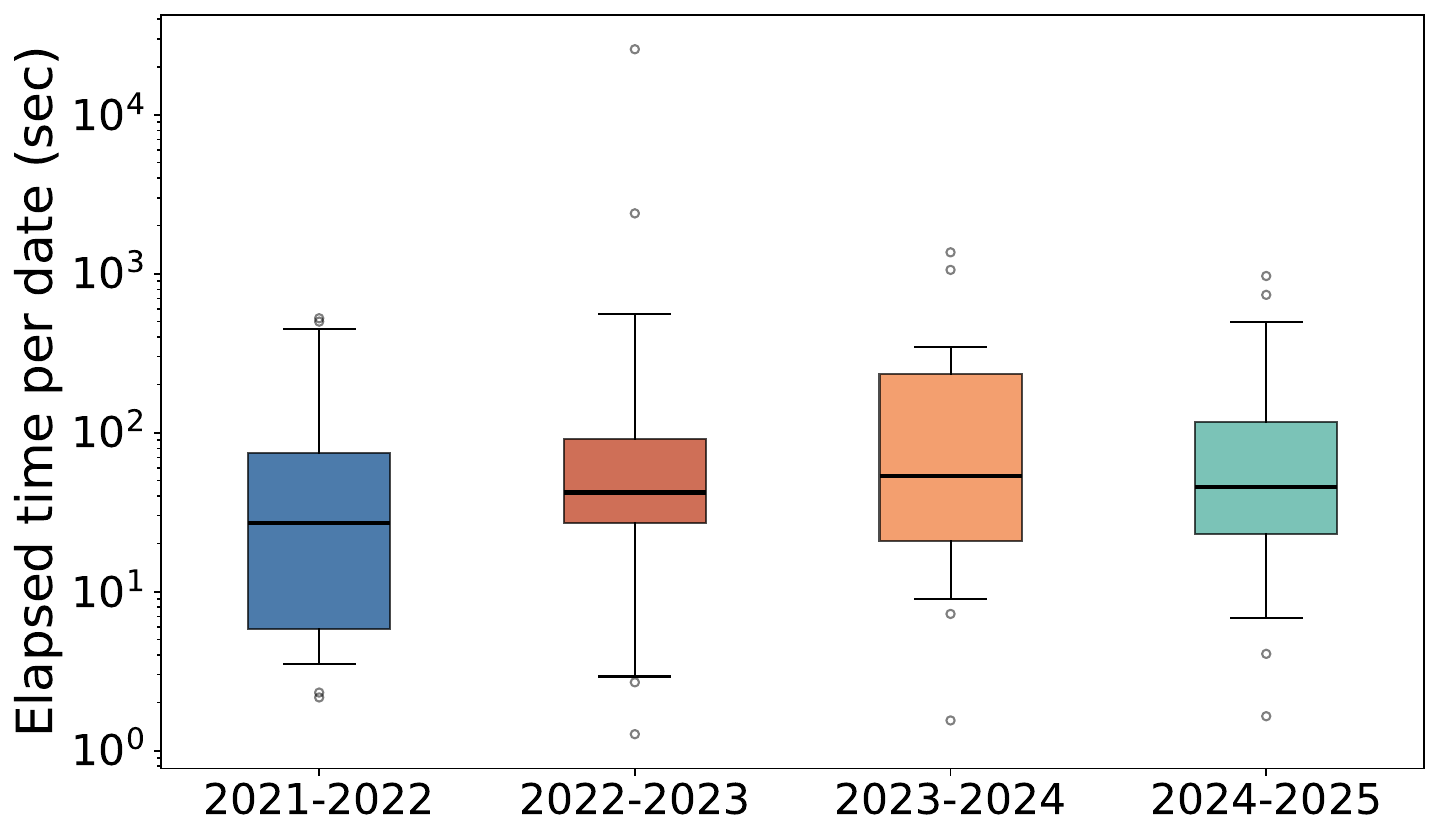}
  \end{subfigure}
  \hfill
  \begin{subfigure}[b]{0.49\textwidth}
    \centering
    \caption{}
    \label{fig:one_day_pruning_efficiency}
    \includegraphics[width=\textwidth]{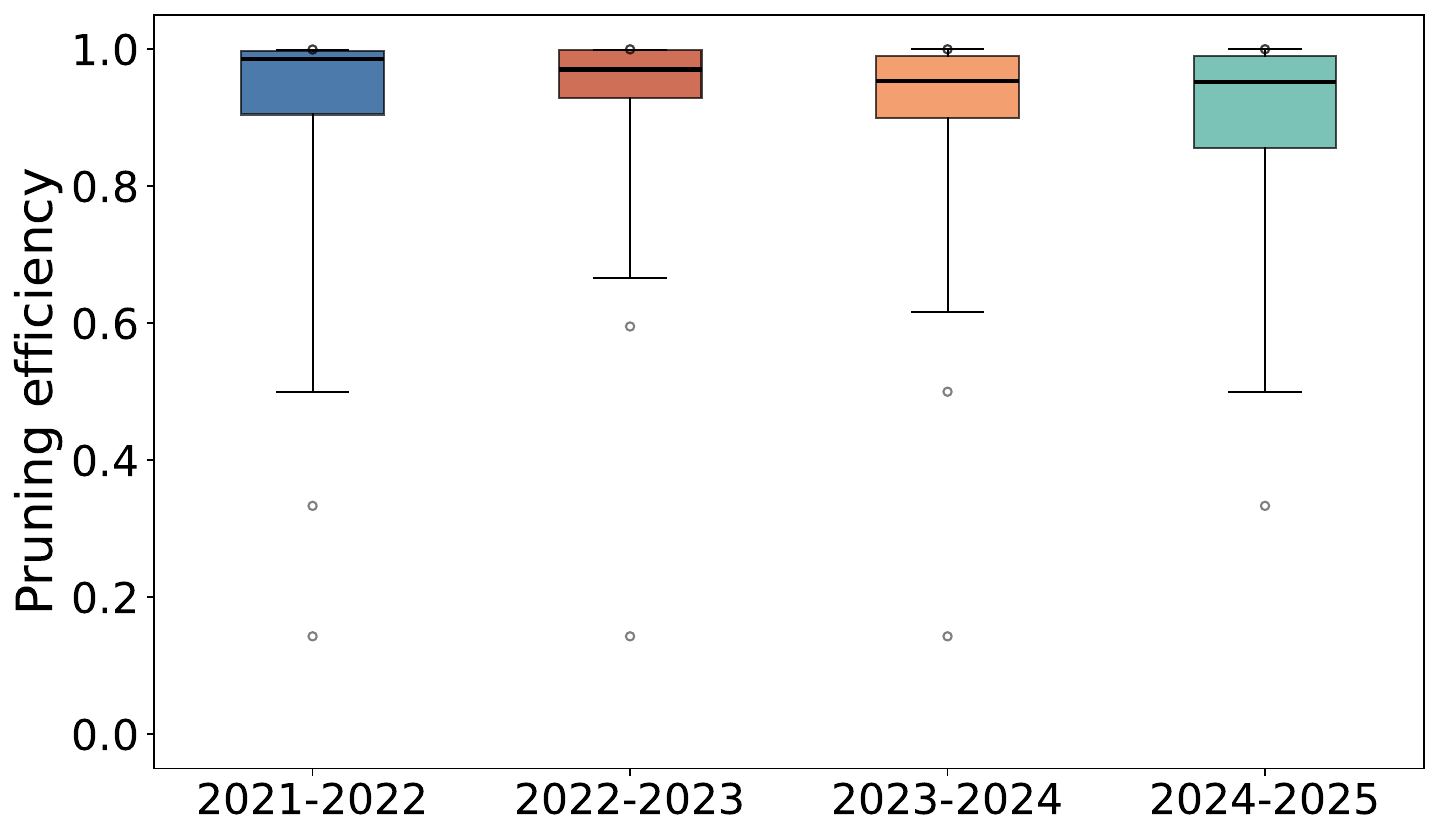}
  \end{subfigure}
  \vspace{-1mm}
  \caption{1-day lookahead approach experiments. \textbf{(a)} Elapsed time to determine clinch scenarios, and \textbf{(b)} Pruning efficiency of the tree search (i.e., the complement of the fraction of nodes explored).}
  \label{fig:one_day}
\end{figure}

\cref{fig:one_day_elapsed_time} shows the statistics of the total time required to determine the 1-day playoff clinching scenarios on each date. The times are fairly consistent across the seasons, however there are several outliers that require significantly more time. The clinching scenarios for the Boston Bruins on March 9th, 2023 had a total literal count of 27, the highest seen in this $n=1$ dataset, as well as the longest elapsed time of 25,759.7 seconds.

\cref{fig:one_day_pruning_efficiency} shows the \emph{pruning efficiency} statistics for the tree search used to determine the 1-day playoff clinching scenarios on each date. We define pruning efficiency as the complement of the fraction of nodes explored out of the total number of nodes in the search tree (considering all six outcomes of all relevant games). For all seasons, most of the pruning efficiencies are clustered near 1, showing the generally high pruning efficiency of our algorithm.

\subsection{Additional Results} 

To develop a better understanding of our methods, we conduct two additional analyses: i) we investigate how often no-goods are generated and/or the goal assignment model is triggered in the 0-day approach, and ii) we test our $n$-day lookahead approach for $n \in \{2,3\}$.

\bigskip
\noindent\textbf{Lazy Constraints and Goal Assignment Model Invocations.} In \cref{tab:solver-stats} we present a summary of the statistics of usage of the goal assignment model and the lazy constraints. From the table it is evident that these mechanisms are only needed in rare cases, however, when needed, they are critical  to make the correct clinching determination. 

\begin{table}[h!]
\centering
\small
\begin{tabular}{l @{\hspace{2em}} r r r r r @{\hspace{2em}} r r r r r}
\hline
 & \multicolumn{5}{c}{0-day} & \multicolumn{5}{c}{1-day} \\
 & & \multicolumn{2}{c}{GA} & \multicolumn{2}{c}{LC} & & \multicolumn{2}{c}{GA} & \multicolumn{2}{c}{LC} \\
Season & N & Total & Max & Total & Max & N & Total & Max & Total & Max \\
\hline
2021-2022 & 708 & 2 & 1 & 3 & 1 & 373 & 9 & 3 & 22 & 14 \\
2022-2023 & 987 & 3 & 1 & 6 & 1 & 739 & 19 & 7 & 36 & 23 \\
2023-2024 & 607 & 9 & 1 & 2 & 1 & 358 & 31 & 7 & 25 & 9 \\
2024-2025 & 830 & 5 & 1 & 4 & 1 & 607 & 11 & 5 & 11 & 2 \\
\hline
\end{tabular}
\vspace{1mm}
\caption{Goal assignment (GA) model and lazy constraint (LC) usage across seasons, for 0-day and 1-day lookahead clinch determination. N is the total number of date-team pairs for which a model was actually built and solved (i.e., excluding trivial cases), Total is the total number of usages of GA/LC (respectively) for that season, and Max is the maximum number of usages of GA/LC (respectively) for a single date-team pair in that season.}
\label{tab:solver-stats}
\end{table}

\bigskip
\noindent\textbf{Longer Lookahead.} Our algorithm can, in principle, be applied to any $n\geq1$ day lookahead. The search space scales like $6^r$ where $r$ is the number of relevant games (as there are six possible game outcomes), and we know the problem is NP-Complete \cite{bernholt1999football, gusfield2002structure, kern2004computational}, so we should expect the elapsed time to scale exponentially. Still, for the 2024-25 season we were able to find all scenarios for $n=1$, almost all for $n=2$ and most of the $n=3$ scenarios with a 30-minute timeout, as shown in \cref{tab:n-day-comparison-timeout}. Instances that take a long time to solve typically result in increasingly complex clinching scenarios; while computationally interesting, these scenarios are less useful to fans and stakeholders as they can be harder to follow. 

\begin{table}[h!]
\centering
\small
\begin{tabular}{r @{\hspace{2em}} r r @{\hspace{2em}} r r @{\hspace{2em}} r r}
\hline
 & & & \multicolumn{2}{l}{Literals} & \multicolumn{2}{c}{Time (s)} \\
$n$ & N & Exc & Max & Med & Max & Med \\
\hline
1 & 40 & 0 & 10 & 2.0 & 901.0 & 17.5 \\
2 & 73 & 3 & 27 & 4.0 & 1381.4 & 44.5 \\
3 & 101 & 35 & 65 & 8.0 & 1769.1 & 87.4 \\
\hline
\end{tabular}
\vspace{1mm}
\caption{Comparison of $n$-day clinching results for the 2024-2025 season with a 30-minute timeout. N is the total number of tree searches, Exc is the number of searches excluded due to timing out, Max and Med are the maximum and median values of the number of literals (in the clinching scenarios) and elapsed time (in seconds), respectively. For $n=2$ running without the timeout resulted in the longest tree search taking 24,735.1s.}
\label{tab:n-day-comparison-timeout}
\end{table}

\section{Conclusion}
\label{sec:conclusion}

We present an algorithm for determining $n$-day playoff clinching
scenarios in the NHL. We build upon previous work by introducing several key
innovations: updated modeling of modern NHL rules and structures, efficient
handling of complex tie-breaking scenarios through a multi-phase approach, and
separation of win assignment and goal assignment models to improve
computational efficiency. The $n$-day algorithm addresses the combinatorial
challenge of multiple future games through intelligent preprocessing to reduce
the search space, along with effective pruning strategies based on game
outcome ordering.

Both the 0-day and the $n$-day lookahead algorithms are tested extensively on public NHL data from the 2021-2022 through
2024-2025 seasons, demonstrating their effectiveness and reliability. The methods
have proven particularly valuable during the latter part of the regular season,
when playoff races intensify and the ability to quickly determine clinching
scenarios becomes crucial for teams, media, and fans. Our implementation pulls game results from a public data source \cite{nhl_api} and produces output matching the NHL's published format.

Future work could focus on extending this algorithm to other clinching scenarios, including
division titles, conference championships, and specific playoff seed achievement. Additionally, the $n$-day lookahead tree search algorithm as well as the 0-day lookahead CP model are fairly general and could be adapted to other sports/leagues with different rules. 

\newpage

\bibliographystyle{plainurl}
\bibliography{references}

@misc{nhl_tiebreakers,
  title = {{NHL} Standings - Tie-Breaking Procedure},
  author = {{National Hockey League}},
  year = {2025},
  url = {https://www.nhl.com/info/standings-info/tie-breaking-procedure},
  note = {Accessed: June 27, 2025}
}

@inproceedings{russell2008mathematically,
  title={Mathematically clinching a playoff spot in the {NHL} and the effect of scoring systems},
  author={Russell, Tyrel and Van Beek, Peter},
  booktitle={Advances in Artificial Intelligence: 21st Conference of the Canadian Society for Computational Studies of Intelligence, Canadian AI 2008 Windsor, Canada, May 28-30, 2008 Proceedings},
  pages={234--245},
  year={2008},
  organization={Springer},
  doi={10.1007/978-3-540-68825-9_23}
}

@book{hooker2011logic,
  title={Logic-based methods for optimization: combining optimization and constraint satisfaction},
  author={Hooker, John},
  year={2011},
  publisher={John Wiley \& Sons}
}

@misc{cpsat,
  title = {{CP-SAT} v9.11.4210},
  author = {Perron, Laurent and Didier, Fr\'{e}d\'{e}ric},
  organization = {Google},
  url = {https://developers.google.com/optimization/cp/cp_solver/},
  year = {2024}
}

@article{kern2004computational,
  title={The computational complexity of the elimination problem in generalized sports competitions},
  author={Kern, Walter and Paulusma, Dani{\"e}l},
  journal={Discrete Optimization},
  volume={1},
  number={2},
  pages={205--214},
  year={2004},
  publisher={Elsevier},
  doi={10.1016/j.disopt.2003.12.003}
}

@phdthesis{russell2010computational,
  title={A computational study of problems in sports},
  author={Russell, Tyrel},
  year={2010},
  school={University of Waterloo}
}

@article{raack2014standings,
  title={Standings in sports competitions using integer programming},
  author={Raack, Christian and Raymond, Annie and Schlechte, Thomas and Werner, Axel},
  journal={Journal of Quantitative Analysis in Sports},
  volume={10},
  number={2},
  pages={131--137},
  year={2014},
  publisher={De Gruyter},
  doi={10.1515/jqas-2013-0111}
}

@article{russell2012hybrid,
  title={A hybrid constraint programming and enumeration approach for solving {NHL} playoff qualification and elimination problems},
  author={Russell, Tyrel and Van Beek, Peter},
  journal={European Journal of Operational Research},
  volume={218},
  number={3},
  pages={819--828},
  year={2012},
  publisher={Elsevier},
  doi={10.1016/j.ejor.2011.11.045}
}

@inproceedings{russell2009determining,
  title={Determining the number of games needed to guarantee an {NHL} playoff spot},
  author={Russell, Tyrel and Van Beek, Peter},
  booktitle={International Conference on Integration of Constraint Programming, Artificial Intelligence, and Operations Research},
  pages={233--247},
  year={2009},
  organization={Springer},
  doi={10.1007/978-3-642-01929-6_18}
}

@article{russell2009lessons,
  title={Lessons learned from modelling the {NHL} playoff qualification problem},
  author={Russell, Tyrel and Van Beek, Peter},
  journal={Constraint Modelling and Reformulation (ModRef'09)},
  pages={132},
  year={2009}
}

@article{wayne2001new,
  title={A new property and a faster algorithm for baseball elimination},
  author={Wayne, Kevin D},
  journal={SIAM Journal on Discrete Mathematics},
  volume={14},
  number={2},
  pages={223--229},
  year={2001},
  publisher={SIAM},
  doi={10.1137/S0895480198348847}
}

@article{robinson1991baseball,
  title={Baseball playoff eliminations: An application of linear programming},
  author={Robinson, Lawrence W},
  journal={Operations Research Letters},
  volume={10},
  number={2},
  pages={67--74},
  year={1991},
  publisher={Elsevier},
  doi={10.1016/0167-6377(91)90089-8}
}

@article{schwartz1966possible,
  title={Possible winners in partially completed tournaments},
  author={Schwartz, Benjamin L},
  journal={SIAM Review},
  volume={8},
  number={3},
  pages={302--308},
  year={1966},
  publisher={SIAM},
  doi={10.1137/1008062}
}

@article{adler2002baseball,
  title={Baseball, optimization, and the world wide web},
  author={Adler, Ilan and Erera, Alan L and Hochbaum, Dorit S and Olinick, Eli V},
  journal={Interfaces},
  volume={32},
  number={2},
  pages={12--22},
  year={2002},
  publisher={INFORMS},
  doi={10.1287/inte.32.2.12.67}
}

@article{ribeiro2005application,
  title={An application of integer programming to playoff elimination in football championships},
  author={Ribeiro, Celso C and Urrutia, Sebasti{\'a}n},
  journal={International Transactions in Operational Research},
  volume={12},
  number={4},
  pages={375--386},
  year={2005},
  publisher={Wiley},
  doi={10.1111/j.1475-3995.2005.00513.x}
}

@article{lucena2008multi,
  title={A multi-agent framework to build integer programming applications to playoff elimination in sports tournaments},
  author={Lucena, Carlos JP and Noronha, Thiago F and Ribeiro, Celso C and Urrutia, Sebastian},
  journal={International Transactions in Operational Research},
  volume={15},
  number={6},
  pages={739--753},
  year={2008},
  publisher={Wiley}
}

@article{donne2012studying,
  title={Studying playoff qualification in motorsports via mixed-integer programming techniques},
  author={{Delle Donne}, Diego and Marenco, Javier},
  journal={Proceedings of the Institution of Mechanical Engineers, Part P: Journal of Sports Engineering and Technology},
  volume={226},
  number={1},
  pages={32--41},
  year={2012},
  publisher={SAGE}
}

@phdthesis{whittle2014nfl,
  title={The {NFL} true fan problem},
  author={Whittle, Scott},
  year={2014},
  school={Kansas State University}
}

@article{gusfield2002structure,
  title={The structure and complexity of sports elimination numbers},
  author={Gusfield, Dan and Martel, Charles},
  journal={Algorithmica},
  volume={32},
  number={1},
  pages={73--86},
  year={2002},
  publisher={Springer},
  doi={10.1007/s00453-001-0074-y}
}

@inproceedings{bernholt1999football,
  title={Football elimination is hard to decide under the 3-point-rule},
  author={Bernholt, Thorsten and G{\"u}lich, Alexander and Hofmeister, Thomas and Schmitt, Niels},
  booktitle={International Symposium on Mathematical Foundations of Computer Science},
  pages={410--418},
  year={1999},
  organization={Springer},
  doi={10.1007/3-540-48340-3_37}
}

@techreport{ito2018calculation,
  title={Calculation of clinch and elimination numbers for sports leagues with multiple tiebreaking criteria},
  author={Ito, Satoshi and Shinano, Yuji},
  institution={Zuse Institute Berlin},
  number={ZIB-Report 18-51},
  year={2018}
}

@phdthesis{husted2019enhancing,
  title={Enhancing Tractability of Mixed-Integer Nonlinear Programming Models: Case Studies in Energy and Sports},
  author={Husted, Mark Andrew},
  year={2019},
  school={Colorado School of Mines}
}

@article{husted2021improving,
  title={Improving sports media's crystal ball for National Basketball Association playoff elimination},
  author={Husted, Mark A and Olinick, Eli V and Newman, Alexandra M},
  journal={INFORMS Journal on Applied Analytics},
  volume={51},
  number={2},
  pages={119--135},
  year={2021},
  publisher={INFORMS},
  doi={10.1287/inte.2020.1034}
}

@article{kim2024improving,
  title={Improving {S}outh {K}orea's Crystal Ball for Baseball Postseason Clinching and Elimination},
  author={Kim, Sam Sung Ho and Husted, Mark A and Olinick, Eli V and Newman, Alexandra M},
  journal={INFORMS Journal on Applied Analytics},
  volume={54},
  number={4},
  pages={298--311},
  year={2024},
  publisher={INFORMS},
  doi={10.1287/inte.2023.0035}
}

@misc{nhl2025clinching,
  author={{NHL}},
  title={Stanley Cup Playoff Clinching Scenarios for {April} 15},
  year={2025},
  url={https://www.nhl.com/news/stanley-cup-playoff-clinching-scenarios-for-april-15-2025},
  urldate={2025-04-15}
}

@misc{nhl_api,
  author={{National Hockey League}},
  title={{NHL Web API}},
  howpublished={\url{https://api-web.nhle.com/}},
  year={2025},
  note={Accessed: 2025-05-05}
}

\end{document}